\documentclass[letterpaper, 10 pt, journal, twoside]{ieeetran}

\IEEEoverridecommandlockouts

\makeatletter
\let\NAT@parse\undefined
\makeatother

\usepackage[caption=false,font=normalsize,labelfont=sf,textfont=sf]{subfig}
\usepackage{textcomp}
\usepackage{stfloats}
\usepackage{url}
\usepackage{verbatim}
\usepackage{catchfilebetweentags}

\usepackage{subcaption}
\usepackage{graphicx}
\usepackage{hyperref}
\usepackage[ruled, linesnumbered, noend]{algorithm2e}
\usepackage{algorithmicx}
\usepackage{amsfonts}
\usepackage{amsmath}
\usepackage[dvipsnames,table,xcdraw]{xcolor}

\newcommand{\Term}[1]{\textsf{#1}}
\newcommand{\func}[1]{\mathtt{#1}}

\usepackage{subcaption}

\usepackage{booktabs}
\usepackage{multirow}
\usepackage[table]{xcolor}
\usepackage{tabularx}
\usepackage{arydshln}
\usepackage{array}
\newcolumntype{C}[1]{>{\centering\arraybackslash}p{#1}}
\newcolumntype{L}[1]{>{\arraybackslash}p{#1}}
\usepackage[shortcuts]{extdash}
\usepackage{cite}

\usepackage{color}

\newcommand{\funcfk}{\func{FK}}
\newcommand{\funcik}{\func{IK}}
\newcommand{\funcvisik}{\func{VisIK}}

\newcommand{\cspace}{C-space}
\newcommand{\cfree}{\ensuremath{\mathcal{C}_{free}}}

\newcommand{\R}{\mathbb{R}}

\newcommand{\visatcfg}[1]{\mathcal{C}\left(#1\right)}

\newcommand{\Set}[2]{\left\{ #1 \;\middle\vert\; #2 \right\}}

\newcommand{\set}[1]{\left\{ {#1} \right\}}

\newcommand{\False}{\Term{FALSE}\xspace}
\newcommand{\True}{\Term{TRUE}\xspace}
\newcommand{\Not}{\Term{NOT}\xspace}
\newcommand{\Or}{\Term{OR}\xspace}

\newcommand{\fov}{\Term{FOV}\xspace}
\newcommand{\tamp}{\Term{TAMP}\xspace}
\newcommand{\vistamp}{\Term{VisTAMP}\xspace}
\newcommand{\tvmp}{\Term{TVMP}\xspace}
\newcommand{\dof}{\Term{DOF}\xspace}

\newcommand{\rrt}{\Term{RRT}\xspace}

\newcommand{\fovrrt}{\Term{FOV-RRT}\xspace}
\newcommand{\smallfovrrt}{{\scriptsize\fovrrt}\xspace}
\newcommand{\prm}{\Term{PRM}\xspace}
\newcommand{\vir}{\Term{VIR}\xspace}

\newcommand{\fovprm}{\Term{FOV-PRM}\xspace}
\newcommand{\smallfovprm}{{\scriptsize\fovprm}\xspace}
\newcommand{\sbmp}{\Term{SBMP}\xspace}

\newcommand{\aabb}{\Term{AABB}\xspace}
\newcommand{\ik}{\Term{IK}\xspace}
\newcommand{\visik}{\Term{VisIK}\xspace}
\newcommand{\fk}{\Term{FK}\xspace}
\newcommand{\vi}{\Term{VI}\xspace}

\newcommand{\new}[1]{\textcolor{red}{#1}}

\graphicspath{{figs/}}

\usepackage{lipsum}

\newcommand{\remove}[1]{}%

\author{Stav Ashur and  Avishai Sintov
\thanks{This work was supported by the Israel Science Foundation (grant No. 451/24).}
\thanks{S. Ashur and  A. Sintov are with the School of Mechanical Engineering, Tel-Aviv University, Israel. Corresponding Author: sintov1@tauex.tau.ac.il.}
}

\title{Sampling-Based Visibility Task Planning}

\begin{document}

\maketitle

\begin{abstract}
Robot Task and Motion Planning (\tamp) algorithms enable autonomous operation by incorporating the specific functions and constraints of end-effector tools, such as grippers or soldering irons, directly into the planning process. In this paper, we explore sampling-based \tamp algorithms specifically designed for a critical subset of devices whose unique properties make traditional planning methods ineffective. Visibility-based instruments, such as exteroceptive sensors, cameras, flashlights and directional antennas, are essential across a vast array of human activities. The unique properties of these devices, and particularly, their field-of-view, render many widely used heuristics and distance metrics less effective. We introduce two new sampling-based algorithms, \fovprm and \fovrrt, designed to tackle visibility-based tasks. \fovprm employs a hierarchical decomposition of the environment, leveraging the concept of \emph{visibility integrity}, to efficiently sample configurations with a clear line-of-sight to the target. A specialized Inverse Kinematics solver enables \fovrrt to ``glance'' in the direction of the target at opportune moments, facilitating the rapid discovery of key configurations. We show that \fovprm and \fovrrt achieve a higher success rate and faster runtimes compared to adaptations of \rrt, \prm and \vir, through both simulated and physical experiments.
\end{abstract}





\section{Introduction}
\label{sec:intro}

\IEEEPARstart{R}{obots} are becoming ever more prevalent in domestic, agricultural, and industrial settings, and perform increasingly complex tasks for which specialized Task and Motion Planning (\tamp) algorithms are required. The common \tamp algorithms often address problems of finding collision-free paths between start and goal configurations and physically manipulating objects \cite{gchkskl-itamp-21}. However, a wide range of applications require establishing a line-of-sight with a target rather than reaching it. Visibility-based \tamp (\vistamp) includes the use of robots equipped with instruments such as cameras, lasers or other directional devices in, for example, visual tracking of objects \cite{ch-vsvt-08}, exploration and mapping \cite{lla-amres-21}, search and search-and-rescue problems \cite{st-vso3deumr-10}
, and UV-powered surface disinfection \cite{mhtls-ubdrr-23}. 

Unlike grippers, drills, or brushes that necessitate physical interaction, visibility-based instruments feature a large, typically elongated, Field-of-View (\fov). This unique geometry challenges the relevance of conventional distance metrics commonly employed in sampling-based planning. The configurations from which a device can successfully observe a target vary widely in both configuration space (\cspace) and workspace distance. Crucially, the target need not even be located within the robot's reachable volume, further complicating the use of traditional proximity-based heuristics that assume the target object is at a reachable location. This makes algorithms and heuristics fitted for robots with short range tools ill-suited for visibility-based instruments.

\begin{figure}
\centering
\begin{tabular}{cc}
   \includegraphics[width=0.42\linewidth]{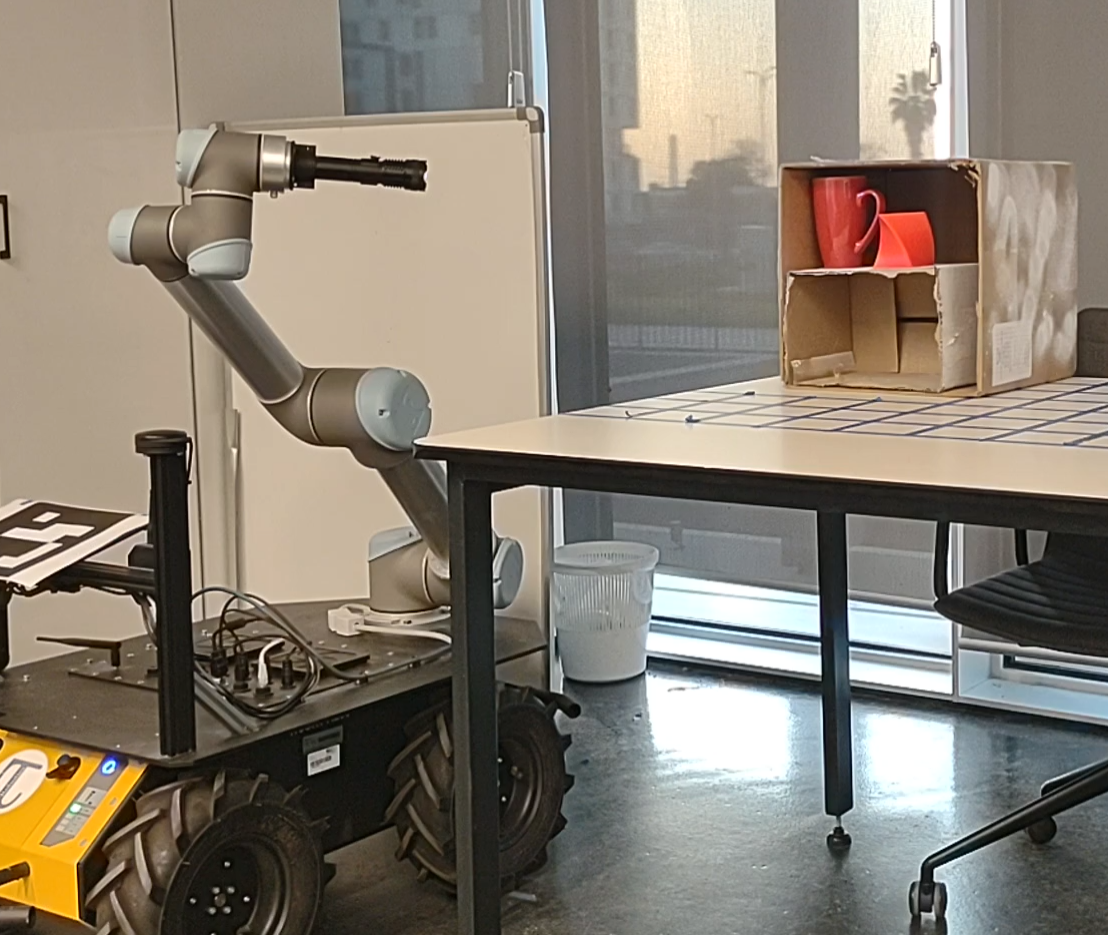} & \includegraphics[width=0.42\linewidth]{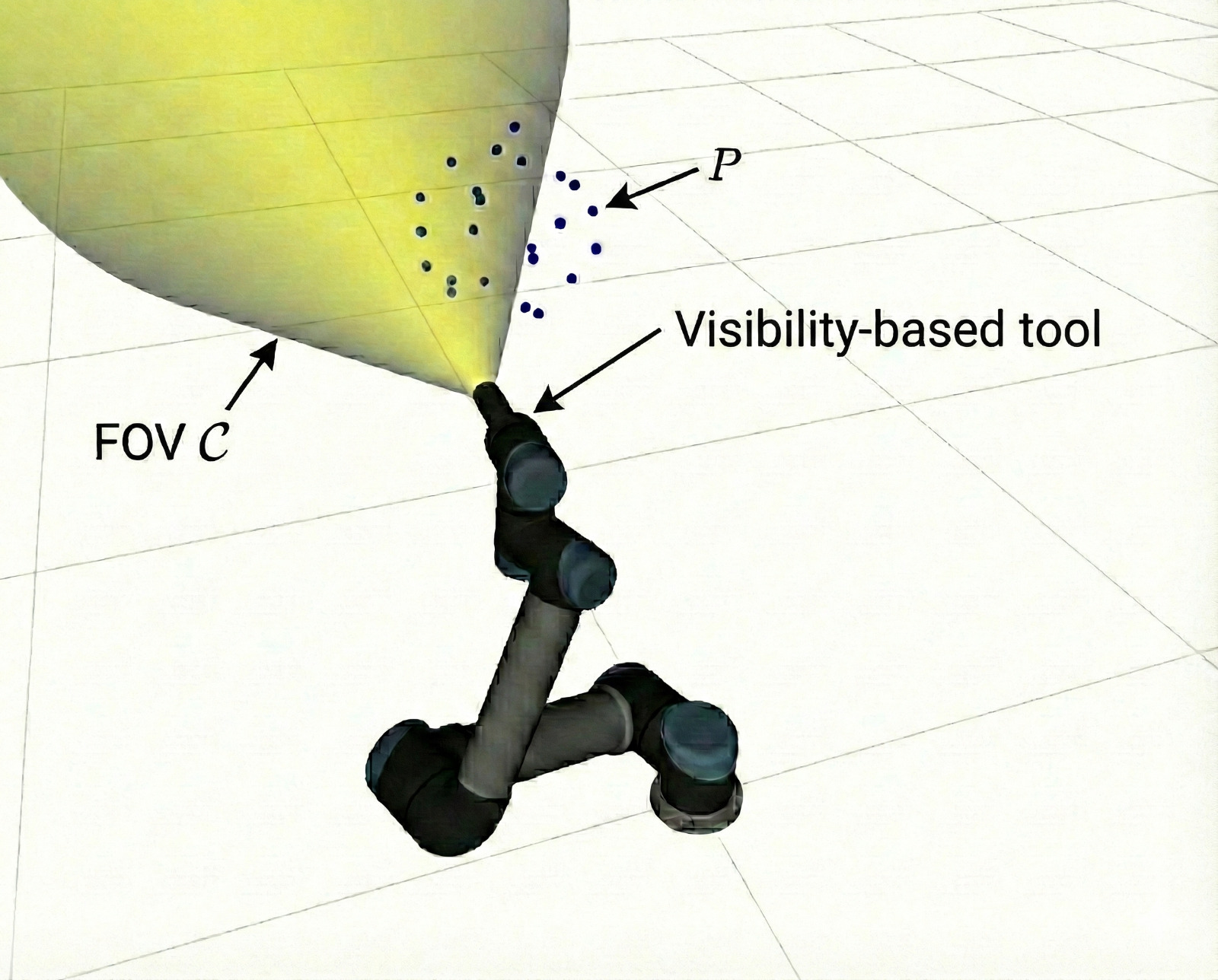} \\
   (a) & (b) 
\end{tabular}
\caption{(a) A robotic arm illuminating two red objects positioned in a shelf. (b) An illustration of a UR5 arm with an attached flashlight partially illuminating a set of points. }
\label{fig:tvmp}
\vspace{-0.5cm}
\end{figure}

Much of the foundational literature focuses on 2D environments where visibility algorithms are abundant and efficient \cite{g-vap-07}, leading to robust solutions for tracking moving targets \cite{lgbl-msmvmt-97} and multi-agent coordination \cite{rl-mrtdtts-16}. However, extending these tasks to 3D space introduces high computational costs and susceptibility to floating-point errors \cite{d-3dvasa-99}. While some spatial approaches utilize Sampling-Based Motion Planning (SBMP) \cite{kslo-prpp-96}, they often rely on simplifying assumptions, such as omni-directional sensor models \cite{smh-sbccrfo3de-05} or unconstrained environments \cite{w-vbopmp-17}. Furthermore, a large body of work on real-world visibility systems, including visual servoing and visual tracking, focuses on inherently local problems \cite{bah-mp3dttao-07, ch-vsvt-08}. These methods typically assume the target is already within the \fov and employ reactive control strategies to maintain visibility \cite{daop-mvdoceummvfi-23}. Such local paradigms are unsuitable for global visibility tasks in constrained environments where the target must first be placed inside the \fov.

While sampling-based planners like \rrt and \prm are well-explored for physical obstacle avoidance, extending them to visibility-driven tasks introduces distinct geometric challenges due to the non-compact, elongated nature of 3D sensor FOVs. This paper bridges this gap by formalizing the Target Visibility Motion Planning (\tvmp), a \vistamp-type problem. \tvmp algorithms provide the essential building blocks required for complex, visibility-dependent tasks (such as 3D visual coverage, visibility-based routing, and multi-robot line-of-sight coordination) where standard planners fundamentally fail to guarantee efficient workspace exploration. We introduce \sbmp algorithms to solve \tvmp in constrained spatial environments, making them applicable across a wide variety of real-world robotic platforms. 

The challenges of \vistamp are addressed through three key contributions. First, we introduce the \emph{Visibility Integrity} (\vi) tree, a novel data structure designed to capture workspace visibility and provide rapid heuristic solutions for complex visibility polyhedron queries. The tree hierarchically partitions the environment into distinct regions, where points within each region maintain approximately uniform visibility characteristics and can be used in the creation of algorithms solving other spatial \vistamp problems. Second, building upon this structure and other building blocks, we present specialized variants of the Probabilistic Roadmap (\prm) \cite{kslo-prpp-96} and Rapidly-exploring Random Tree (\rrt) \cite{l-rrtnt-98} algorithms, termed Field-of-View \prm (\fovprm) and Field-of-View \rrt (\fovrrt), respectively. Finally, we provide a comprehensive evaluation of these algorithms in both simulated and real-world environments. Our experimental results, including a physical demonstration with a robot-mounted flashlight (Figure \ref{fig:tvmp}a), reveal significant performance gains in runtime and success rate over \prm \cite{mrph-ocpuvsd-21}, \rrt, and Visibility Integrity Roadmap (\vir) \cite{zhvml-c3dfrvuvi-19} benchmarks adapted for \tvmp. The planner code is open-source\footnote{\url{https://github.com/StavAshur/visibility_task_planning}} for potential benchmarks and to advance research in the field.

\section{Related work}
\label{sec:related:work}

\paragraph*{Planar visibility task planning}
Much of the existing literature on \vistamp focuses on planar environments, where efficient and, in some cases, optimal solutions often exist. Planar \vistamp algorithms benefit from the simplicity and abundance of visibility algorithms \cite{g-vap-07}. Optimal algorithms for achieving and maintaining visibility of a single moving target with known or partially-predictable movements were introduced in \cite{lgbl-msmvmt-97}. Unpredictable moving targets were later addressed in \cite{MCTH-sbmpamvut-05}. Many more planar \vistamp problems were tackled, such as landmark-based navigation \cite{bmh-oplbnddvfvc-07}, simultaneous tracking and localization \cite{wlzhlf-ovmpfttl-14}, pursuit-evasion games \cite{bh-enetppegvc-08}, and multi-agent tracking \cite{rl-mrtdtts-16}.

\paragraph*{Spatial visibility task planning} 
Spatial \vistamp requires solving spatial visibility problems, some of which are computationally expensive and susceptible to floating point errors \cite{d-3dvasa-99}. To plan trajectories for multi-target tracking, a \new{Visibility Integrity Roadmap} (\vir) based on \vi was established in \cite{lk-fmtvw-16, zhvml-c3dfrvuvi-19}. \vir assumes a 3D point robot with an omni-directional sensor, answering queries by sampling within high-\vi clusters, derived from either a grid or randomly sampled points. Conversely, our method supports complex systems by using a hierarchical decomposition to compute visibility relationships between different regions. 
Another \prm variant was used to compute efficient paths for exhaustively searching an object in a 3D environment \cite{smh-sbccrfo3de-05}. 
Similarly, a \prm graph was augmented to produce motion plans maintaining a target in the robot's \fov \cite{blcl-ppivuprm-10}. A different work uses \prm for surface disinfection using UV light \cite{mrph-ocpuvsd-21}. It constructs a roadmap while keeping track of the regions illuminated by the UV device in the sampled configurations. The roadmap is then used to plan a path covering the set of target surfaces. We compare our proposed algorithms to baseline \prm and \rrt planners based on this work. A modified $\text{RRT}^*$ framework was deployed in \cite{byam-lprrtdsrccf-22} using various local planners to optimize motion paths across distinct cost functions. However, their visibility application is strictly limited to maintaining a target within the FOV, bypassing the problem of active target exploration. Another work plans paths that maintain visibility with a target or avoid entering a sentry's visibility region \cite{pld-prpmpsvr-24}.

\paragraph*{Other spatial visibility tasks}
A large body of work is dedicated to local visibility tasks. Visual servoing \cite{ch-vsvt-08} is a widely-used technique for robot motion planning and control using a camera. Visual servoing problems include the eye-in-hand model in which a camera is attached to the robot and motion planning is performed where the goal state is a configuration in which the camera's view matches an input image. In the visual tracking problem \cite{bah-mp3dttao-07,ghnd-kmmtfvmc-11,lwlcz-udmavtcrehfcc-15,daop-mvdoceummvfi-23}, an adjustable camera or a camera-wielding robot must keep a moving target or set of targets within its \fov. While these problems share similarities with the \tvmp, they typically assume an initial state where the target is already within the \fov. Consequently, they focus on local reactive strategies to maintain visibility, making control decisions based on the target’s relative displacement and immediate workspace constraints. Algorithms developed for such tasks are thus unsuitable for global visibility tasks, and cannot be used as benchmarks in our experiments. Another set of \vistamp algorithms are designed only for controlled conditions such as robots in unconstrained environments or with a predefined set of allowable paths \cite{w-vbopmp-17}, e.g., on rails.

\paragraph*{Sampling-based motion planning}
The proposed algorithms, \fovrrt and \fovprm, follow the \sbmp paradigm in which a geometric graph $G=(V,E)$ is constructed in the robot's \cspace ~using random sampling. $V$ and $E$ are associated with robot configurations and continuous motions respectively, and an $(s,t)$-path in $G$ is a motion plan from configuration $s$ to $t$. \sbmp emerged with the introduction of the \prm algorithm \cite{kslo-prpp-96}. This approach utilizes a roadmap data structure to approximate the connectivity and topology of the free space (\cfree), the set of all valid configurations in \cspace. Motion planning queries are answered by connecting the input start and goal configurations to the roadmap. \rrt \cite{l-rrtnt-98,kl-rrtceasqpp-00} is a single-query algorithm that constructs a search tree in \cspace ~by extending new edges from graph nodes toward random samples. In contrast to real-world physical visibility, \cite{sln-vbprmp-00} utilizes \cspace{} visibility to create sparse roadmaps, rejecting samples in \cspace{} regions already visible by existing nodes. \sbmp algorithms were found to be effective for a wide range of motion planning problems, and many more variants of \rrt and \prm as well as new \sbmp algorithms were added over the years. See \cite{ock-smpcr-24} for a comparative review of \sbmp algorithms.

\section{Preliminaries}
\label{sec:preliminaries}

\subsection{Problem definition}
We formally define the \tvmp problem below. While this is, to the best of our knowledge, the first time the problem has been explicitly characterized in literature, it is a fundamental challenge inherent to many real-world applications.

Let $r$ be a robot equipped with a visibility-based tool characterized by a conic (or conic frustum) \fov beam $\mathcal{C}$, and let $\visatcfg{x}$ denote the conic \fov emitted from $r$ at configuration $x$. Given an environment (workspace) $E \subseteq \R^3$, a target set $P\subseteq E$, and an initial configuration $s$ of $r$, the goal is to compute a motion plan that transitions $r$ from $s$ to some terminal configuration $t$, such that $P\subseteq \visatcfg{t}$. An illustration is given in Figure \ref{fig:tvmp}b. A natural variant of this problem, termed $\alpha$-\tvmp, requires that at least an $\alpha$-fraction of $P$ is contained within $\visatcfg{t}$, for some $\alpha \in (0, 1)$. Furthermore, this definition may extend to an optimization problem not included in the scope of this work: finding a configuration $t$ that maximizes the visible fraction of $P$.

\subsection{Simplifying assumptions}
In this work, we adopt several simplifying assumptions detailed below along with their underlying justifications. 

\emph{\textbf{Target set.}}
While the target set $P$ can be any arbitrary set of points in $E$, our algorithms first compute an approximation of the minimum enclosing sphere $P_S$ of $P$, and solve the \tvmp problem for $P_S$ rather than $P$. The reasoning behind this assumption is based on several considerations. 
\begin{enumerate}
    \item The most likely use-case for this algorithm may involve a single object or a compact region in $\R^3$.
    \item The assumption remains valid in the case of multiple target objects not occluded by obstacles.
    \item The case where the target objects are occluded by obstacles or are placed in close proximity to obstacles, e.g., in different compartments in a shelf or on a table, can be handled by solving an $\alpha$-\tvmp problem.
\end{enumerate}
We further assume that the points within $P$ are non-occluding. Given that $P$ is modeled as a sphere, this implies that all viewpoints yielding the same visible surface area are functionally equivalent. Consequently, the planning task does not require a specific orientation relative to the target.

\emph{\textbf{Environment.}}
We assume that the environment can be sufficiently approximated by basic geometries such as Axis-Aligned Bounding Boxes (\aabb), spheres, or triangular meshes. This assumption is required to leverage GPU-accelerated operations for processing large batches of visibility queries efficiently. Such approximations of environmental and robotic geometry are standard practice in motion planning literature to ensure computational tractability \cite{tkk-mmvsbp-23}.

\subsection{Visibility integrity (\vi)}
\begin{figure}[!t]
\centering
\subfloat[]{\includegraphics[width=0.4\linewidth,page=14]{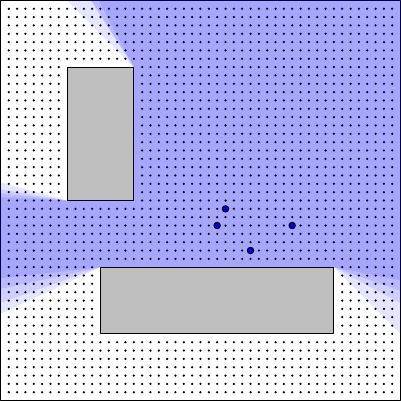}%
\label{fig:vi:example:high}}
\hfil
\subfloat[]{\includegraphics[width=0.4\linewidth,page=15]{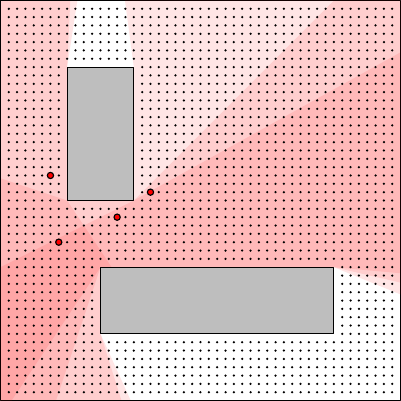}%
\label{fig:vi:example:low}}
\caption{Visibility examples in a planar environment represented by a set of grid points $X$ (black). (a) The blue points exhibit high \vi, as they share nearly identical visibility of the points in $X$. (b) The red points have divergent visibility of points in $X$ yielding a low \vi.}
\label{fig:vi:example}
\end{figure}

To efficiently handle visibility queries, \fovprm maintains a tree data structure based on the \vi measure \cite{lk-fmtvw-16, zhvml-c3dfrvuvi-19}. The \vi score of a point set is an approximation of the intersection-over-union of their visibility polyhedrons. Thus, \vi measures the degree to which a set of points shares a common visible workspace (see Figure \ref{fig:vi:example}). Given a finite point set $X\subseteq \R^d$, we denote the subset of $X$ visible from a point $p\in \R^d$ by $V_X(p)\subseteq X$. The \vi of a point set $P\subseteq \R^d$ is defined by
\begin{equation}
    VI(P) = \frac{|\Set{\bigcap V_X(p)}{p\in P}|}{|\Set{\bigcup V_X(p)}{p\in P}|},
\end{equation}
where operator $|\cdot |$ denotes the cardinality of a discrete set.

\section{Method}
\label{sec:method}

In this section, we present the proposed \vistamp algorithms \fovprm and \fovrrt. We begin by introducing two algorithmic components that enable \prm and \rrt to address the \tvmp problem. Specifically, we describe the \vi-tree and the visibility \ik solver in Section \ref{sec:method:adjusments}. The pseudocode and a detailed explanation of \fovprm and \fovrrt are given in Sections \ref{sec:method:fovprm} and \ref{sec:method:fovrrt}, respectively. Assuming familiarity with the standard \prm and \rrt algorithms, we focus on the modifications that distinguish \fovprm and \fovrrt from their traditional counterparts.

\subsection{Adjustments to \sbmp algorithms}
\label{sec:method:adjusments}

The primary challenge in adapting \prm and \rrt for \tvmp is the absence of a predefined goal configuration. Unlike the motion planning problem where both start and goal configurations are explicitly provided as input, the \tvmp query provides only a start configuration and a set of target points $P$ that must be observed. To address the challenge of identifying suitable goal configurations, we introduce two algorithmic components. First, the \vi-tree guides \fovprm in sampling goal regions with high visibility potential. Second, a visibility-aware \ik solver leverages seed configurations to generate goals for \fovrrt extensions.

\subsubsection{Visibility integrity tree}
\label{sec:method:visibility:integrity:tree}

Identifying a valid goal configuration mandates reasoning over workspace visibility, as the implicit goal set corresponds to the intersection of visibility polyhedrons associated with the points of $P_S$. Formally, the objective is to find a set of goal configurations $\mathcal{T}$ such that every $t\in \mathcal{T}$ satisfies $P_S \subset \visatcfg{t}$. Hence, we require a computational mechanism that can efficiently identify workspace regions from which $P_S$ is visible. Computing 3D visibility polyhedrons is expensive, and exact algorithms are prone to numerical errors \cite{d-3dvasa-99}. We propose a tree-based data structure that partitions the environment into regions characterized by high \vi scores and captures their mutual visibility relationships. Then, given a query $P_S$, regions with good visibility of $P_S$ can quickly be located.  

Pseudocode of the tree building process is given in Algorithm \ref{alg:vi:tree}. The root of the tree is assigned with an \aabb containing the entire workspace $E$ (Line \ref{vi:tree:line:root}). A recursive tree-building function is called in Line \ref{vi:tree:line:recursive:tree}. Below we describe the recursive function,after which we continue to the second part of Algorithm \ref{alg:vi:tree} in which visibility relationships between regions are inferred.

\begin{figure}

\centering
\subfloat[]{\includegraphics[trim={0mm 10mm 0mm 10mm}, clip, width=0.48\linewidth]{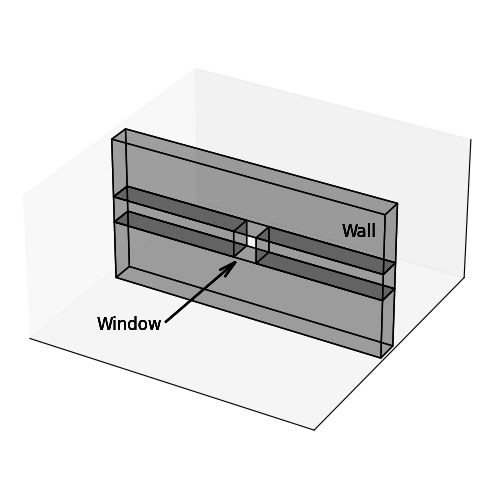}%
\label{fig_env_a}}
\hfil
\subfloat[]{\includegraphics[trim={0mm 10mm 0mm 10mm}, clip, width=0.48\linewidth]{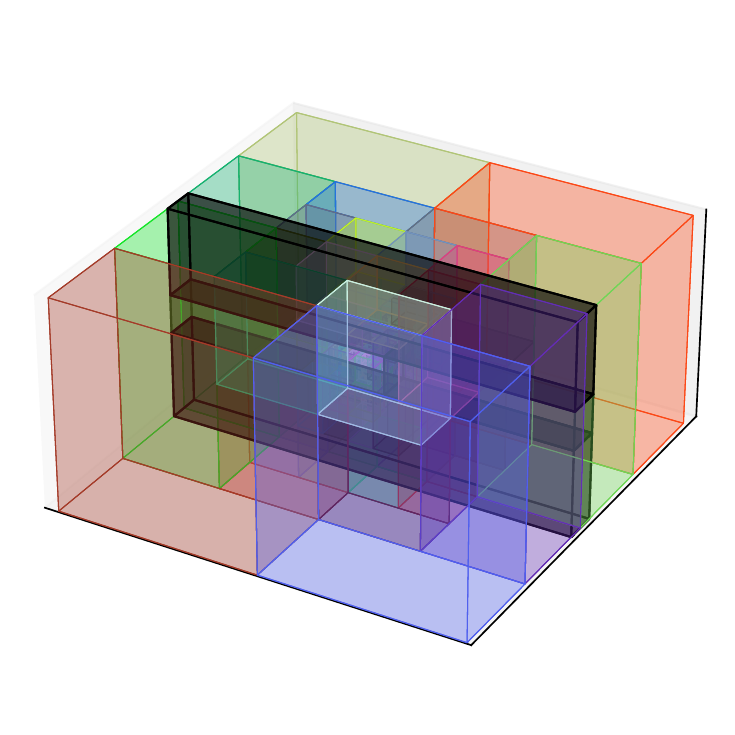}%
\label{fig_env_b}}
\caption{(a) An example environment of a wall with a window, and (b) the corresponding leaf \aabb{}s of the \vi-tree.}
\label{fig:svi:leaves}
\end{figure}

The construction of the \vi-tree, detailed in Algorithm \ref{alg:vi:tree:recursive}, recursively generates the subtree for an input root node. First, samples are taken from within the root's \aabb (Line \ref{vi:tree:recursive:line:ample:in}) and from its complement (Line \ref{vi:tree:recursive:line:ample:out}). This sample partitioning reflects our application of \vi in establishing global visibility relationships between disjoint regions. The outward-\vi computation (Line \ref{vi:tree:recursive:line:vi}) serves as the termination criterion for the recursion. If the resulting score satisfies the leaf-node condition, the algorithm concludes for that branch. Otherwise, the current \aabb is split (Line \ref{vi:tree:recursive:line:split}), and the procedure recursively initializes two child nodes. See Figure \ref{fig:svi:leaves} for a visualization of the leaf \aabb{}s in a \vi-tree.

Returning to Algorithm \ref{alg:vi:tree}, after the tree structure is finalized the algorithm performs a bottom-up process of populating the nodes' visibility lists, containing the information regarding mutual visibility in the workspace. Starting at the leaves (Line \ref{vi:tree:line:leaf:visibility}), the centers of every pair of leaves $\set{l_1,l_2}$ are tested for mutual visibility, and, if successful, they are added to each other's lists. The minimum outward-\vi score of the leaves ensures that point-to-point visibility generalizes to the majority of the pairs in $l_1.aabb \times l_2.aabb$. Following leaf processing, the algorithm performs a post-order traversal (Line \ref{vi:tree:line:bottom:up}) to propagate visibility information upward. Each internal node's visibility list is defined as the intersection of its children's lists. Simultaneously, the node is inserted into the visibility lists of those appearing in the intersection. This completes the tree construction.

\begin{algorithm}
    \caption{Visibility-integrity tree construction}%
    \label{alg:vi:tree}
    \SetAlgoLined%
    \SetKwInOut{Input}{input}
    \SetKwInOut{Output}{output}

    \Input{Environment $E$, threshold $\tau$}

    \tcp{\color{OliveGreen}\# Building \vi tree}
    $root.aabb \gets E.\func{get\_bounding\_box}()$ \Comment{get AABB of the entire environment}
    \label{vi:tree:line:root}\\
    $T\gets \func{build\_tree\_recursive}(root, \tau)$
    \label{vi:tree:line:recursive:tree} \Comment{Alg. \ref{alg:vi:tree:recursive}}
    
    \tcp{\color{OliveGreen}\# Computing visibility bottom-up}
    $P \gets \Set{l.aabb.center}{ l \in T.leaves}$
    \Comment{leaf centers}\\
    $M_{vis} \gets \func{pairwise\_visibility}(P)$
    \label{vi:tree:line:pairwise:leaf:visibility}\\
    \ForEach{$\set{l_1,l_2}\in \binom{T.leaves}{2}$ \label{vi:tree:line:leaf:visibility}}{
        \If{$M_{vis}[l_1.idx][l_2.idx] = 1$}{
            $l_1.visibility\_list.\func{insert}(l_2.idx)$\\
            $l_2.visibility\_list.\func{insert}(l_1.idx)$\\
        }
    }
    \ForEach{$v \in \func{post\_order(T)}$\label{vi:tree:line:bottom:up}}{
        $seen\_by\_children \gets v.left.visibility\_list \cap v.right.visibility\_list$\\
        $v.visibility\_list \gets seen\_by\_children$\\
        \ForEach{$u \in seen\_by\_children$}{
            $T.nodes[u].visibility\_list.\func{insert}(v.idx)$
        }
    }
   \Return
\end{algorithm}

\begin{algorithm}[b]
    \caption{Recursive construction of the VI-tree}%
    \label{alg:vi:tree:recursive}
    \SetAlgoLined%
    \SetKwInOut{Input}{input}
    \SetKwInOut{Output}{output}
    \Input{Node $root$, threshold $\tau$}
    $b \gets root.aabb$\\
    $S_{in} \gets \func{sample\_uniform}(b, n_{samples})$
    \label{vi:tree:recursive:line:ample:in}\\
    $S_{out} \gets \func{sample\_uniform}(E.\func{get\_bounding\_box}() \setminus b , n_{samples})$
    \label{vi:tree:recursive:line:ample:out}
    \Comment{sample outside \aabb}\\
    $score \gets \frac{|\func{seen\_by\_all}(S_{in},S_{out})|}{|\func{seen\_by\_any}(S_{in},S_{out})|}$ \Comment{\vi computation}
    \label{vi:tree:recursive:line:vi}\\
    \If{$score > \tau$}{
        $root.is\_leaf \gets \True$ \Comment{this node is a leaf}\\
        $T.leaves.\func{insert}(root)$\\
        \Return
    }\Else{
        \tcp{\# create children and recurse}
        $b_{l}, b_{r} \gets \func{split\_aabb}(b)$
        \label{vi:tree:recursive:line:split}
        \Comment{split root's \aabb}\\
        $root.left \gets \func{build\_tree\_recursive}(root, b_{l})$\\
        $root.right \gets \func{build\_tree\_recursive}(root, b_{r})$
    }
    \Return  
\end{algorithm}

\paragraph{Query} \label{par:vi:tree:query} The result of a query to the \vi-tree, whose pseudocode is presented in Algorithm \ref{alg:vi:tree:query}, is a map (dictionary) with $(node, volume)$ pairs that associate between nodes of the tree and an estimated lower bound on the fraction of the target's volume they see. The query is recursive, and begins at the root of the tree. When visiting a node $v$, the base case is checked. If $v.aabb$ is contained in $P_S$ or is a leaf, the result list is updated with $v$'s visibility list and, for each member of the list, an estimated fraction of $v.aabb \cap P_S $ it sees (Line \ref{vi:tree:query:line:base:intersect}). Otherwise, recursive calls to the children are made in Line \ref{vi:tree:query:line:recurse}, assuming their \aabb{}s intersect $P_S$.

\subsubsection{Visibility Inverse Kinematics function}
\label{sec:method:vis:ik}

Standard \ik computations return a configuration that places the robot's end-effector at a specific position or pose. In \vistamp settings, we require an \ik map (i.e., $\ik : E \longrightarrow \mathcal{C}_{space}$) that outputs a configuration $c$ such that $P_S \subset \visatcfg{c}$. Hence, our visibility-based \ik solver (\visik) does not receive an input pose, but must instead derive a suitable pose from the input target $P_S$. To focus this optimization, \visik relies on a seed configuration $g$, which acts as a geometric hint representing a valid robot posture whose end-effector has an unobstructed line-of-sight to the target. This allows \visik to restrict its search to configurations pointing directly at the target. This dependency on a seed arises from the inherently global nature of the \fov’s role as an end-effector, where small changes in \cspace{} or $E$ can lead to significant shifts in the projected visibility. Consequently, we attempt to produce meaningful output configurations by requiring that for the seed configuration $g$, $P_S \cap \visatcfg{g} > \beta$ holds; that is, the position of the end-effector at configuration $g$ sees a minimum fraction of the target. Under this assumption, we can increase the value of returned configurations without explicitly reasoning over visibility.

\begin{algorithm}
    \caption{Recursive VI-tree query}%
    \label{alg:vi:tree:query}
    \SetAlgoLined%
    \SetKwInOut{Input}{input}
    \SetKwInOut{Output}{output}
    \Input{Node $root$, target set $P_S$, result map $res$}
    \tcp{\color{OliveGreen}\# base case - contained or leaf}
    \If{$root.aabb \subseteq P_S ~\Or~ root.is\_leaf $ \label{vi:tree:query:line:base:intersect}} {
        \ForEach{$v \in root.visibility\_list$}{
            $normalized\_volume \gets \frac{\func{volume}(root.aabb \cap P_S)}{\func{volume}(P_S)}$
            \Comment{visibility value of seeing $root$}
            \label{vi:tree:query:line:volume}\\
            \If{$v \in res.keys$}{
                $res[v] \gets res[v] + normalized\_volume$
            }
            \Else{
                $res.\func{insert}(v, normalized\_volume)$
            }
        }
        \Return
    }
    \tcp{\color{OliveGreen}\# recursive calls}
    \If{$root.left.aabb \cap P_S \neq \emptyset$ \label{vi:tree:query:line:recurse}} {
        $\func{recursive\_vi\_query}(root.left, P_S, res)$
    }
    \If{$root.right.aabb \cap P_S \neq \emptyset$} {
        $\func{recursive\_vi\_query}(root.right, P_S, res)$
    }
    \Return
\end{algorithm}

Using the simplifying assumption converting $P$ into the sphere $P_S$, the algorithm computes a workspace line segment $\overline{s}$ which is contained in the line defined by the position of the end-effector and the center of $P_s$. Hence, any point $x\in \overline{s}$ has the same visibility of $P_S$ as the end-effector's position of $r$ at the seed configuration $g$. This is done by searching along the line connecting $p=\fk(g)$ to the center of $P_S$, where function $\fk:\mathcal{C}_{space} \longrightarrow E$ is the position Forward Kinematics (\fk) of $r$. The segment $\overline{s}$ is further intersected with the $r$'s reachable volume to rule out unreachable poses.

Once the segment $\overline{s}$ is finalized, it is discretized into a series of positions. Each position is lifted to a full $SE(3)$ pose by aligning the "forward" axis with the center of $P_S$. A standard \ik solver can then process these intermediate poses to generate a sequence of configurations. Although computationally expensive, we perform these \ik calls immediately to enable a \cspace{} nearest-neighbor search between the seed configuration and the intermediates. Any \ik solver can be used as a black box for this purpose, making this subroutine easy to implement and optimize. The algorithm then evaluates these configurations in order of increasing distance from the seed, returning the first valid result. This prioritization of \cspace ~proximity is particularly beneficial in \fovrrt (Algorithm \ref{alg:vis:rrt}). Given a valid seed $g$, a proximal configuration is more likely to share that validity, thereby increasing the probability of a successful tree extension.


\subsection{Field-of-View \prm}
\label{sec:method:fovprm}

Algorithm \ref{alg:vis:prm} contains the pseudocode of \fovprm, a modified \prm algorithm capable of solving multi-query \tvmp. To maintain brevity, we present the construction and query phases within a unified algorithm, formulated around a single query $(s, P_S)$ as the primary input argument. The algorithm starts by constructing the two data structures that serve as the backbone of the algorithm: A \cspace{} roadmap graph $G$ is constructed in Line \ref{prm:line:c:space:prm} as in the standard \prm algorithm, and a \vi-tree in Line \ref{prm:line:vi:tree}. 
The latter is capable of sampling configurations likely to be viable goal states from which the query point set $P_S$ is visible.

Next, the algorithm connects the start configuration $s$ to $G$, choosing neighbors by some distance metric. Our implementation utilizes Euclidean \cspace{} distance, though other distance metrics are also applicable. The first attempt to find valid goal configurations follows immediately in Line \ref{prm:line:roadmap:coverage}, where the \fov{}s of configurations stored in $G$ are exhaustively tested. While this subproblem easily admits a geometric data structure, we found the time required to scan the roadmap nodes is far from being the algorithm's bottleneck, assuming that the \fov is saved at every graph node. If the roadmap or the connected component of $s$ do not yet contain a valid goal state, two steps are executed. The graph is extended with new configuration nodes (Line \ref{prm:line:roadmap:sample}), and goal configurations are sampled (Line \ref{prm:line:find:vis:neighborhood}) by lifting the output of \vi-tree queries to $SE(3)$ and applying \ik computations. The former is an attempt to find more connecting opportunities and merge existing connected components. The latter addresses the core \tvmp challenge of determining goal configurations from the target set.

Once both $s$ and a member $t$ of the goal configuration set $X$ have been connected via $G$, an $(s,t)$-path is returned using any Single Source Shortest Path (SSSP) algorithm.

\subsection{Field-of-View-\rrt (\fovrrt)}
\label{sec:method:fovrrt}

Algorithm \ref{alg:vis:rrt} contains the pseudocode of \fovrrt, a modified \rrt algorithm for instances of single query \tvmp. The algorithm extends the tree in a normal \rrt manner, finding the nearest neighbor based on Euclidean \cspace{} distance and maintaining the crucial Voronoi bias (Line \ref{rrt:line:extend}). While standard \rrt often biases the search by sampling the goal configuration in a fraction of iterations, this approach is inapplicable to \fovrrt as no explicit goal configuration is available. Moreover, integrating the \vi-tree used by \fovprm would introduce a heavy pre-processing phase to an otherwise agile algorithm, a cost that, unlike in the \fovprm framework, cannot be amortized across multiple tasks.

In the standard \rrt, if an endpoint of a successful extension is sufficiently close to the goal configuration, a connection toward that goal is attempted. Similarly, \fovrrt employs a visibility heuristic as an oracle to identify promising nodes with a high probability of seeing the target. Upon identification, the \visik solver leverages the node to generate a feasible goal configuration toward which the tree can extend. Each time an extension operation results in the addition of a configuration $c'$ to the tree (Line \ref{rrt:line:extend}), the algorithm invokes $\fk(c')$ to compute the position $p$ of the visibility device, i.e., the apex of the \fov{} (Line \ref{rrt:line:fk}), and performs ray shooting queries to points of $P_S$ (Line \ref{rrt:line:is:close}). When the count of successful visibility queries exceeds a predefined threshold $n_{visibility}$, the algorithm invokes the \visik solver (Line \ref{rrt:line:vis:ik}) to derive a nearby goal configuration. Once found, a direct connection to this newly generated goal state is attempted (Line \ref{rrt:line:vis:extend_visik}).

\begin{algorithm}
    \caption{Field-of-View \prm}%
    \label{alg:vis:prm}
    \SetAlgoLined%
    \SetKwInOut{Input}{input}
    \Input{$r, E, s, P_S$}
    $G \gets \func{PRM}(r, E, n_{samples})$
    \label{prm:line:c:space:prm}
    \Comment{build roadmap for $r$ in $E$}\\
    $T \gets \func{visibility\_integrity\_tree(E)}$
    \label{prm:line:vi:tree}
    \Comment{Alg. \ref{alg:vi:tree}} \\

    $connected\_start\_cfg \gets \func{connect}(s,G)$\Comment{connect based on \cspace{} distance} \label{prm:line:connect} \\
    $X \gets \emptyset$  \Comment{initializing set of targets}\\
    $connected\_goal\_cfg \gets \False$\\
    \ForEach{$c \in G$ \label{prm:line:roadmap:coverage}}{
        \If{$P_S \subseteq \visatcfg{c}$}{
            $connected\_goal\_cfg \gets G.\func{is\_connected(s,c)}$\\
            $X.\func{insert}(c)$\\    
        }
    }
    \While{(\Not $connected\_goal\_cfg$ ) \Or (\Not $connected\_start\_cfg$)}{
        $H \gets \func{sample\_c\_space}(r, E, n_{batch})$ \label{prm:line:roadmap:sample}\Comment{sample new configurations}\\
        $\func{connect}(H,G)$ \Comment{same as Line \ref{prm:line:connect} for the set $H$}\\
        \If{\Not $connected\_start\_cfg$}{
            $connected\_start\_cfg \gets\func{connect}(s,G)$\\
        }
        $N_t^{\R^3} \gets T.\func{sample\_goal\_positions}(P_S)$\Comment{use \vi-tree to sample from $E$. See Sec. \ref{par:vi:tree:query}}\label{prm:line:find:vis:neighborhood} \\
        $N_t^{SE(3)}\gets \func{points\_to\_poses}(N_t^{\R^3}, P_S)$
        \Comment{lift samples to full 6\dof poses} \\     
        $X.\func{insert}(\funcik(N_t))$ 
        \Comment{store goal configurations}\\
        $connected\_goal\_cfg \gets G.\func{is\_connected}(s, X)$\\    
    }
    $\Return$ $\func{SSSP}(G,s,X)$ \Comment{find shortest path to a goal}
\end{algorithm}

\paragraph*{\textbf{A note on probabilistic completeness}} 
Let $\mathcal{A} \subseteq \mathcal{C}_{\text{space}}$ denote the subset of configurations in \cspace{} that successfully observe the target. If $\mathcal{A}$ has a non-zero measure and is reachable from $s$ via paths with minimum clearance $\delta > 0$, then both proposed algorithms are probabilistically complete,
as a direct corollary of Theorem 1 in \cite{kslbh-pcrrtgkpfp-19}.

\begin{algorithm}[b]
    \caption{Field-of-Viewbased \rrt}%
    \label{alg:vis:rrt}
    \SetAlgoLined%
    \SetKwInOut{Input}{input}
    \Input{$r, E, s, P_S$}
    $T \gets ({s}, \emptyset$) \Comment{initializing an empty tree}\\
    $found\_goal\_cfg \gets \False$\\

    \While{\Not $found\_goal\_cfg$}{
        $c\gets \func{sample\_c\_space}(r, E)$ \label{rrt:line:sample:c:space} \Comment{random sampling}\\
        $c' \gets \func{extend}(T, c)$ \Comment{standard \rrt extension}
        \label{rrt:line:extend}\\
            $p \gets \funcfk{}(c')$ \Comment{forward kinemtaics computation} 
            \label{rrt:line:fk}\\
            $is\_close \gets \left(\func{visibility}(p, P_S)  > n_{visibility}\right)$\Comment{Check if the target is visible from the end-effector position sample} \label{rrt:line:is:close}\\
            \If{$is\_close$}{
                $t \gets \funcvisik{}(c', P_S)$\Comment{finds a configuration close to $c'$ that sees $P_S$}\label{rrt:line:vis:ik}\\
                $\func{extend}(T, t)$\label{rrt:line:vis:extend_visik}\\
            }
    }
    $\Return$ $\func{SSSP}(T,s,t)$
\end{algorithm}


\section{Experiments}
\label{sec:experiments}

In this section, we describe in detail the experimental setup and the experiments conducted to compare our methods to baselines. Our experiments include a large number of simulated experiments to validate the algorithms, and a small number of real world experiments. In the simulated experiments, we use a 6-dof UR5 mounted on a 3-\dof planar mobile platform, yielding in a 9-\dof robotic system. In the physical experiments, we use a fixed-base 6-\dof UR5, see Fig. \ref{fig:physical:experiments:two:2}a. Although the physical arm is mounted on a mobile base, the platform was kept stationary throughout the real-world experiments. A flashlight attached to the arm's end-effector served as our visibility-based instrument. We assign a range of $h=3$~m to the flashlight, and aperture angles of $\gamma\approx30^\circ$ and $\gamma\approx10^\circ$ in the simulated and real-world experiments, respectively.%

To ensure consistency across evaluations, targets were pre\-/instantiated and shared across all algorithms. Internal algorithmic stochasticity, e.g., sampling subroutines, remained independent. Since randomly generated instances $(r, E, s, P_S)$ do not guarantee the existence of an admissible solution, all reported runtimes correspond exclusively to successful trials unless otherwise stated. In all of the experiments planning was done on a desktop PC with an Intel i7-14700F CPU, 64 GB of RAM, and a GeForce RTX 4090 GPU. A video illustrating the simulated and real-world experiments is included in the supplementary materials.


\subsection{Algorithms evaluation}
\label{sec:experiments:sim}

\begin{figure}
\centering
\begin{tabular}{cc}
    \includegraphics[height=3.2cm]{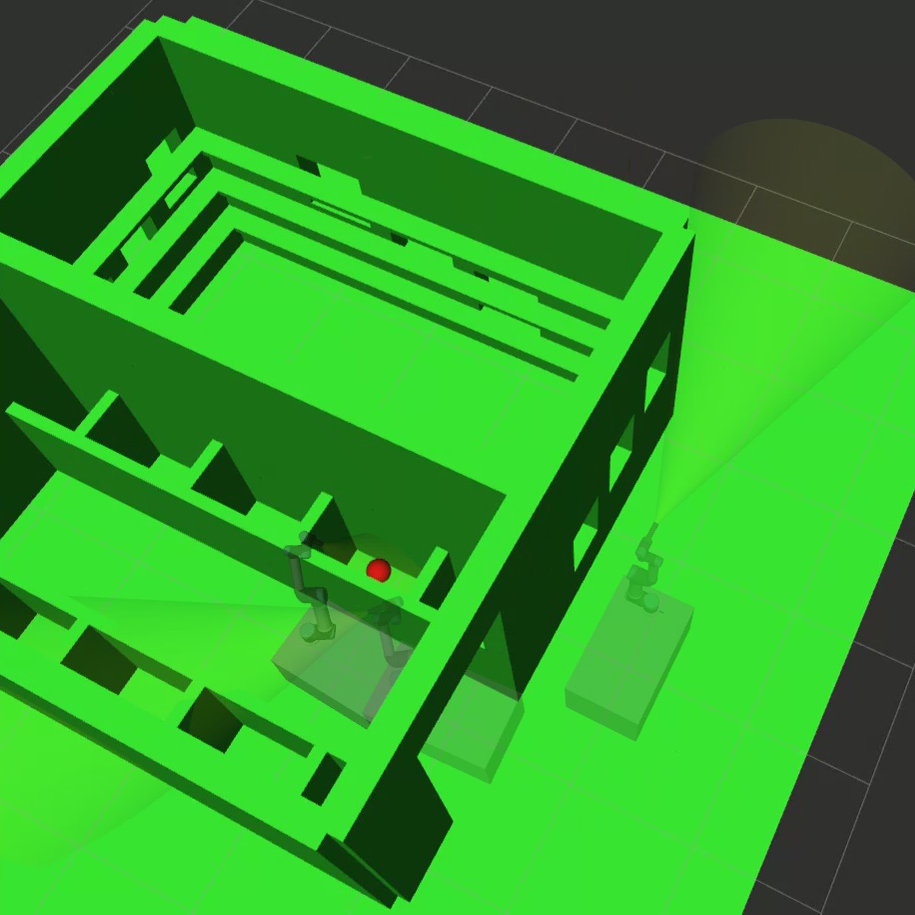} &  \includegraphics[height=3.2cm]{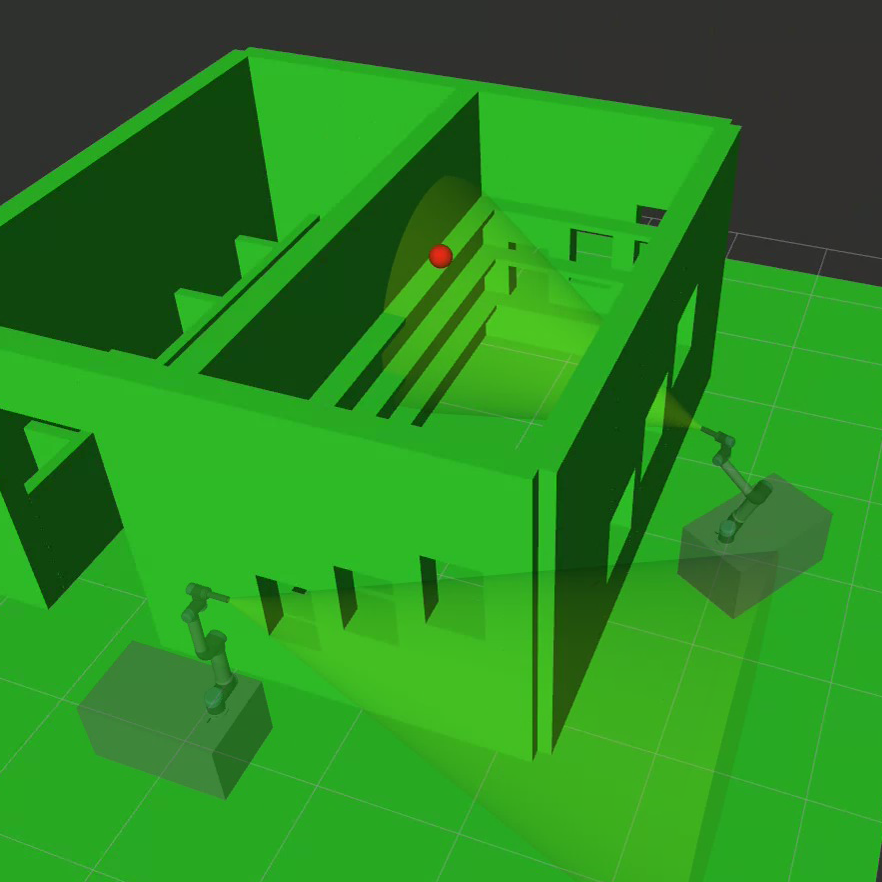}\\

\end{tabular}
\caption{Simulated (left) door and (right) windows environments with a 9-DOF robot moving to illuminate a sphere.}
\label{fig:simulated:experiments}
\end{figure}

The proposed VisTAMP algorithms were evaluated in the simulation shown in Figure \ref{fig:simulated:experiments}. The environment comprises two central rooms encircled by a wide corridor. The first room is highly constrained, featuring a narrow passage flanked by ten deep, open-top containers (five per side) that restrict base rotation and severely limit the configurations capable of achieving line-of-sight to the container floors. The second room is an unconstrained workspace with three windows per outer wall and three shelves per inner wall. Being obstacle-free, its valid configurations form a large connected component in \cspace. Spherical targets, whose radii match their clearance to the nearest obstacle, were randomly sampled within the containers or on the shelves.

We evaluate the performance of \fovprm and \fovrrt by comparing them against \tvmp adaptations of \prm, \rrt, and an adaptation of the \vir algorithm. The baseline \prm modification is drawn from \cite{mrph-ocpuvsd-21}, which addresses a surface-covering problem for UV disinfection. In this framework, a roadmap $G=(V,E)$ is constructed using uniform sampling, and given a query $(s,P_S)$, $s$ is connected to $G$, and $V$ is scanned for a node $v$ such that $P_S$ is sufficiently covered by $\visatcfg{v}$. If no such configuration is found, the roadmap is expanded, and the \fov of configurations is evaluated as possible goals as they are added. In the case of the baseline \rrt, we simply check the \fov of every new configuration added, to determine if it satisfies the visibility requirements for $P_S$. Because these baselines represent minimal adaptations for the \tvmp problem, we retain their names to avoid introducing redundant nomenclature. The original \vir algorithm \cite{zhvml-c3dfrvuvi-19} assumes a simple point robot and cannot be applied directly to complex manipulators. To adapt it as a benchmark, we have modified the query structure, evaluate cluster visibility relationships, and use retrieved points to compute robot configurations. We use its random sampling variant, as the grid approach yields preprocessing runtimes orders of magnitude worse than our proposed method.

Each of the five tested algorithms was invoked for 200 trials with a time limit of 120 seconds each. The \rrt variants, \rrt and \fovrrt, were given a fixed start configuration and no information was saved between runs of the experiment. \prm variants, \fovprm, \vir, and \prm itself, were given randomized valid start configurations, excluding configurations inside door-less room in order to assure task feasibility, and maintained the aggregated roadmap throughout the entire experiment. Runtime statistics and success rates were recorded, and are presented in Table \ref{tab:sim:exp}. The results show a clear success rate and runtime advantage to \fovrrt and \fovprm over the compared algorithms. The separation of the two tasks demonstrates that the narrow passage in the door room favors the \prm variants, excluding the benchmark \prm. We attribute \prm's poor performance to the small probability of sampling a manipulator configuration illuminating inside the containers.

\begin{figure}[h]
    \centering
    \includegraphics[width=\linewidth]{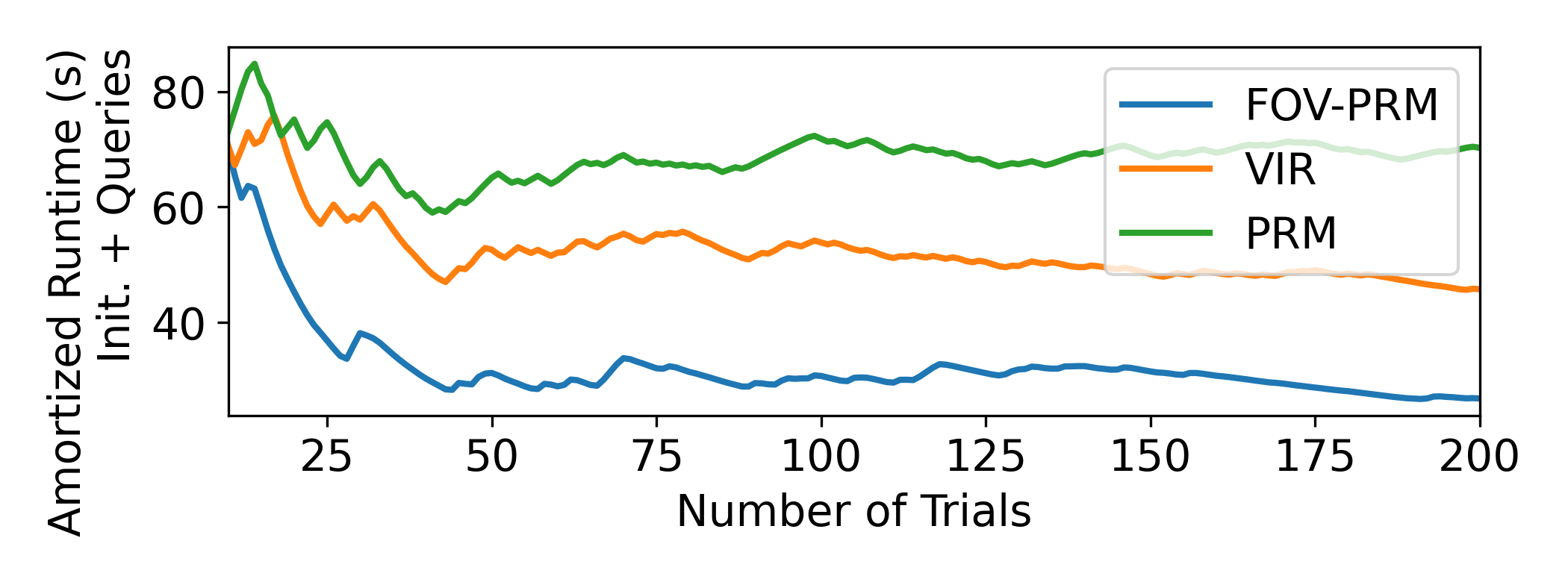}
    \caption{Amortized runtime across 200 successful and failed trials, including \fovprm, \vir  and \prm overhead. Each point denotes the cumulative runtime over consecutive trials.
    }
    \label{fig:results:amortized}
\end{figure}

Figure \ref{fig:results:amortized} shows the improvement in runtime of \prm variants as the number of queries increases. These trends suggest that the aggregated statistics in Table \ref{tab:sim:exp} obscure the superior performance of \fovprm, which becomes significantly more pronounced as the roadmap grows. Note that the runtime average includes the pre-processing time for building the \vi-tree and \vir's workspace clustering. For \prm, the minimal runtime deviation indicates that success is mostly limited to queries immediately answered by the current roadmap.


\subsection{Real-world demonstrations}

\begin{figure}[h]
    \centering
    \begin{tabular}{cc}
        \includegraphics[height=3cm]{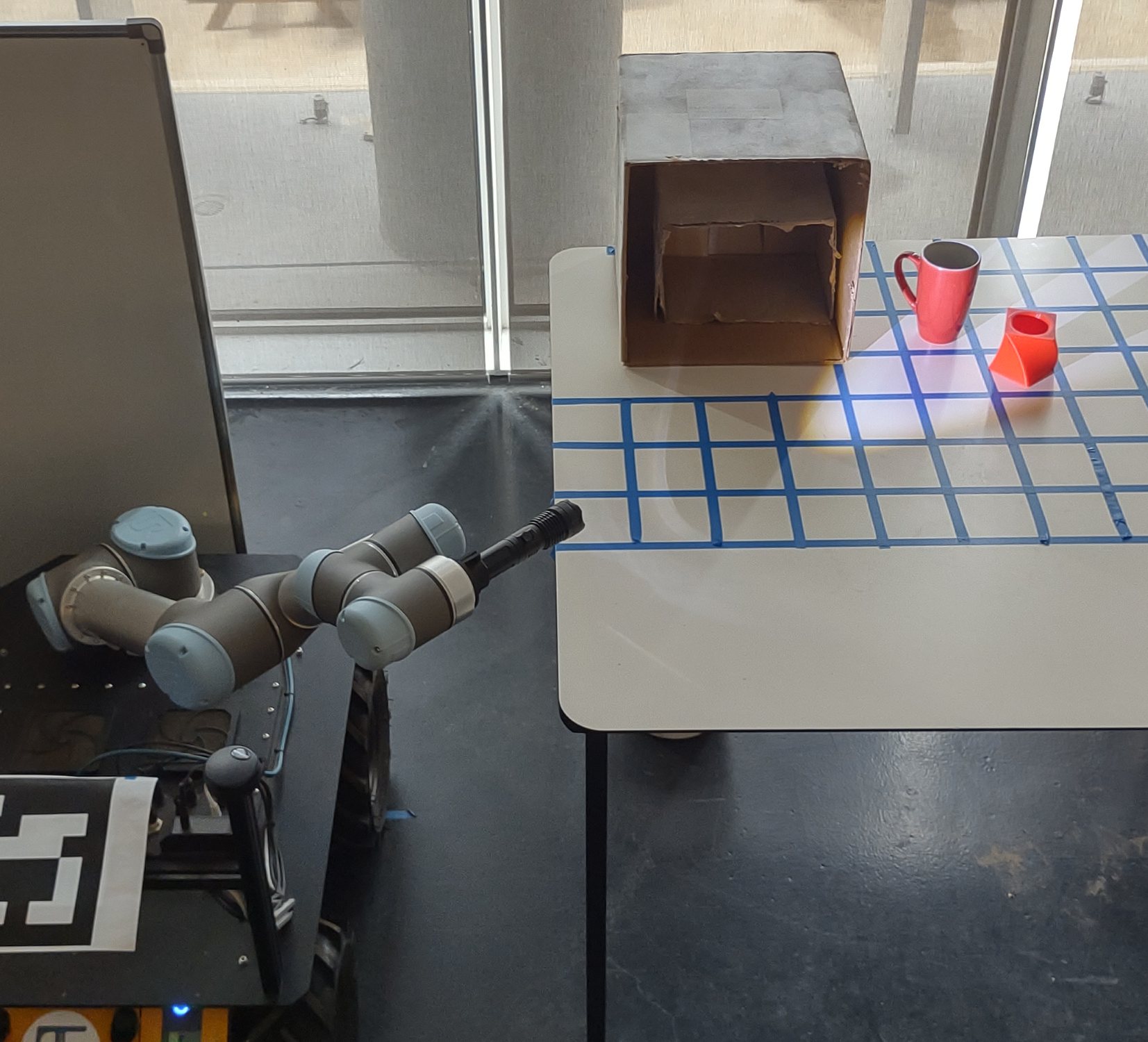} & \includegraphics[height=3cm]{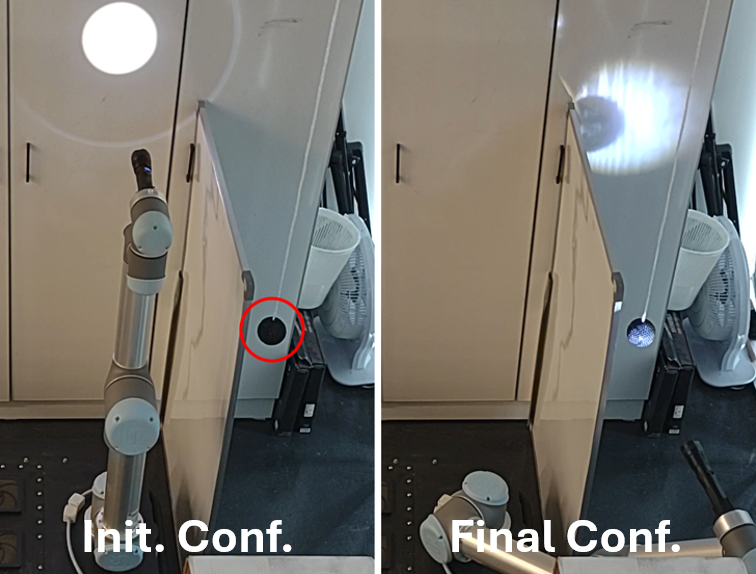} \\
        (a) & (b)
    \end{tabular}
    \caption{Experiments of (a) a manipulator illuminating two red targets within a constrained workspace, and (b) a test scenario requiring illumination of a wall-occluded ball.}
    \label{fig:physical:experiments:two:2}
\end{figure}

The proposed \tvmp algorithms were further validated on a physical UR5 manipulator. The robotic arm was situated in a constrained workspace featuring a table with a three-tiered shelving unit constructed from nested boxes. A coordinate grid was overlaid on the table and each shelf level to facilitate consistent positioning. The experimental task required the illumination of two target objects, a mug and a pen holder, placed at random locations. Examples of robot and objects positioning can be seen in Figures \ref{fig:tvmp}b and \ref{fig:physical:experiments:two:2}. A ZED 2 depth camera mounted on the ceiling tracked the 3D coordinates of these objects in real-time. Before each trial, the algorithm computed a bounding sphere encompassing both targets to define the visibility target for the \tvmp solver. 

\begin{table}
    \centering
    \caption{Performance results for the simulated experiments of the window and door rooms}.

    \begin{tabular}{L{1.13cm} C{0.3cm} C{0.5cm} C{0.3cm} C{1.7cm} C{0.4cm} C{0.4cm} C{0.6cm}}
        \toprule

        \multirow{2}{*}{Algorithm} & \multicolumn{3}{c}{Success rate (\%)} & \multicolumn{4}{c}{Comp. time (s)} \\
        \cmidrule(lr){2-4} \cmidrule(lr){5-8} 
        & All & Win. & Door & Mean & Med. & Min. & Max. \\
        \midrule
        {\smallfovrrt} & 64.0 & 90.4 & 29.1 & 6.61 $\pm$ 12.56  & 1.48 & 0.09 & 68.21 \\
        \rrt    & 37.5 & 62.3 & 4.7  & 14.94 $\pm$ 21.61 & 5.82 & 0.09 & 119.26 \\
        
        \midrule
        
        {\smallfovprm} & 87.0 & 87.7 & 86.0 & 9.12 $\pm$ 15.15 & 2.48 & 0.11 & 94.81 \\
        \vir     & 69.0 & 75.4 & 60.5 & 9.80 $\pm$ 15.84  & 4.30 & 0.25 & 97.23 \\
        \prm    & 44.0 & 77.2 & 0.0  & 4.97 $\pm$ 9.53  & 1.61 & 0.28 & 71.78 \\
        \bottomrule
    \end{tabular}
    \label{tab:sim:exp}
\end{table}

While the environment constrained the robot's motions, the ZED 2 setup prohibited creating more occlusion. To compensate and increase task difficulty, we restricted the flashlight’s \fov to a narrow beam of $\gamma=10^\circ$ and enforced a 60-second time limit per query. We compared all four algorithms under these settings, while also including a $30^\circ$ \fov variant of \prm as an additional baseline. Each algorithm was evaluated over 20 trials. The \rrt variants were initialized from a fixed starting configuration. In contrast, the \prm variants planned from the robot’s current state toward the target sets, with roadmaps incrementally aggregated throughout the duration of the experiment.

Query success required generating a valid path and physically executing it to illuminate both target objects within a strict 60-second time limit, with results summarized in Table \ref{tab:physical:exp}. Despite the highly constrained 60-second budget and a narrow $10^\circ$ \fov{} beam, \fovprm and \fovrrt significantly outperformed the baseline benchmarks, resolving approximately half of all queries. The baseline \prm devoted more computational overhead to nearest-neighbor searches and edge validation, adding fewer nodes and examining fewer \fov{}s than \rrt. Furthermore, the spatial concentration of targets on the table surface favored \fovprm and \fovrrt, as configurations sampled by the \vi-tree to observe one target were highly reusable for others. Finally, to match the coverage efficiency inherently provided by our visibility-aware sampling, the baseline \prm required increasing the FOV angle to $\gamma=30^\circ$—a threefold increase in angle yielding a roughly 9-times larger \fov area. Another demonstration of illuminating an occluded ball target is shown in Figure \ref{fig:physical:experiments:two:2}b and in the supplementary video.

\begin{table}
    \centering
    \caption{Success rates for real robot experiments}
    \footnotesize
    \setlength{\tabcolsep}{8pt} 
    \begin{tabular}{lccccc}
        \toprule
        Algorithm & \prm & \prm & \fovprm & \rrt & \fovrrt \\
        & ($10^\circ$) & ($30^\circ$) & & & \\
        \midrule
        Suc. Rate (\%) & 0 & 45 & 50 & 10 & 40 \\
        \bottomrule
    \end{tabular}
    \label{tab:physical:exp}
\end{table}


\section{Conclusion}
\label{sec:conclusion}
We have defined the \tvmp problem, a core \vistamp task with numerous real-world applications, and introduced two \sbmp algorithms solving it, \fovprm and \fovrrt. Together with the algorithms, we have also developed a \vi-based tree and an \ik solver for visibility devices, enabling the adaptation of \sbmp methods to \vistamp settings. The algorithms were evaluated against \sbmp benchmarks in both simulated and real-world environments, and were shown to achieve higher success rates and faster runtimes.

\emph{\textbf{Future work.}} First, we aim to deploy the mobile manipulator and test our methods in unstructured, real-world environments. Also, we intend to explore the applicability of the \sbmp paradigm for \vistamp tasks such as coverage, search, and pursuit-evasion. Additionally, we strive to integrate general learning-based visibility oracles in \vistamp algorithms to enhance their applicability. 

\vspace{-0.1cm}


\bibliographystyle{IEEEtran}
\bibliography{robotics.bib}

@INPROCEEDINGS{tkk-mmvsbp-23,
    author={Thomason, Wil and Kingston, Zachary and Kavraki, Lydia E.},
  booktitle={{IEEE} Int. Conf. on Rob. \& Auto.}, 
  title={Motions in Microseconds via Vectorized Sampling-Based Planning}, 
  year={2024},
  volume={},
  number={},
  pages={8749-8756},
}

@inproceedings{lgbl-msmvmt-97,
  author       = {Steven M. LaValle and
                  H{\'{e}}ctor H. Gonz{\'{a}}lez{-}Ba{\~{n}}os and
                  Craig Becker and
                  Jean{-}Claude Latombe},
  title        = {Motion strategies for maintaining visibility of a moving target},
  booktitle    = {{IEEE} International Conference on Robotics},
  pages        = {731--736},
  year         = {1997},
}

@incollection{ch-vsvt-08,
  author       = {Fran{\c{c}}ois Chaumette and
                  Seth Hutchinson},
  title        = {Visual Servoing and Visual Tracking},
  booktitle    = {Springer Handbook of Robotics},
  pages        = {563--583},
  publisher    = {Springer},
  year         = {2008},
}

@article{blcl-ppivuprm-10,
  author       = {Matthew A. Baumann and
                  Simon L{\'{e}}onard and
                  Elizabeth A. Croft and
                  James J. Little},
  title        = {Path Planning for Improved Visibility Using a Probabilistic Road Map},
  journal      = {{IEEE} Trans. Robotics},
  volume       = {26},
  number       = {1},
  pages        = {195--200},
  year         = {2010},
}

@article{mhtls-ubdrr-23,
  title={{UV} disinfection robots: a review},
  author={Mehta, Ishaan and Hsueh, Hao-Ya and Taghipour, Sharareh and Li, Wenbin and Saeedi, Sajad},
  journal={Rob. Aut. Sys.},
  volume={161},
  year={2023},
}

@article{lla-amres-21,
  author       = {Iker Lluvia and
                  Elena Lazkano and
                  Ander Ansuategi},
  title        = {Active Mapping and Robot Exploration: {A} Survey},
  journal      = {Sensors},
  volume       = {21},
  number       = {7},
  pages        = {2445},
  year         = {2021},
}

@article{st-vso3deumr-10,
  author       = {Ksenia Shubina and
                  John K. Tsotsos},
  title        = {Visual search for an object in a 3D environment using a mobile robot},
  journal      = {Comput. Vis. Image Underst.},
  volume       = {114},
  number       = {5},
  pages        = {535--547},
  year         = {2010},
}

@book{g-vap-07,
place={Cambridge},
title={Visibility Algorithms in the Plane},
publisher={Cambridge University Press},
author={Ghosh, Subir Kumar},
year={2007}
}

@article{lwlcz-udmavtcrehfcc-15,
  author       = {Xinwu Liang and
                  Hesheng Wang and
                  Yunhui Liu and
                  Weidong Chen and
                  Jie Zhao},
  title        = {A unified design method for adaptive visual tracking control of robots
                  with eye-in-hand/fixed camera configuration},
  journal      = {Autom.},
  volume       = {59},
  pages        = {97--105},
  year         = {2015},
}

@inproceedings{daop-mvdoceummvfi-23,
  author       = {Fatih Dursun and
                  Bruno Vilhena Adorno and
                  Simon Watson and
                  Wei Pan},
  title        = {Maintaining Visibility of Dynamic Objects in Cluttered Environments Using Mobile Manipulators and Vector Field Inequalities},
  booktitle    = {{IEEE/RSJ} Int. Conf. on Intel. Rob.},
  pages        = {6371--6378},
  year         = {2023},
}

@phdthesis{d-3dvasa-99,
  author       = {Fr{\'{e}}do Durand},
  title        = {Visibilit{\'{e}} tridimensionnelle : {\'{e}}tude analytique
                  et apllications. {(3D} Visibility : Analytical Study and Applications)},
  school       = {Joseph Fourier University, Grenoble, France},
  year         = {1999},
}

@inproceedings{mrph-ocpuvsd-21,
  author       = {Jo{\~{a}}o Marcos Correia Marques and
                  Ramya Ramalingam and
                  Zherong Pan and
                  Kris Hauser},
  title        = {Optimized Coverage Planning for {UV} Surface Disinfection},
  booktitle    = {{IEEE} Int. Conf. on Rob. \& Auto.},
  pages        = {9731--9737},
  year         = {2021},
}

@inproceedings{pld-prpmpsvr-24,
    author = {John Phillips and Sihui Li and Neil T. Dantam},
    booktitle = {Algorithmic Foundations of Robotics XVI (WAFR)},
    title = {Park Rangers' Problem: Motion Planning for Sequential Visibility Requirements},
    year = {2024}
}

@article{zhvml-c3dfrvuvi-19,
  author       = {Jixuan Zhi and
                  Yue Hao and
                  Christopher Vo and
                  Marco Morales and
                  Jyh{-}Ming Lien},
  title        = {Computing 3-D From-Region Visibility Using Visibility Integrity},
  journal      = {{IEEE} Robotics Autom. Lett.},
  volume       = {4},
  number       = {4},
  pages        = {4286--4291},
  year         = {2019},
}

@inproceedings{wlzhlf-ovmpfttl-14,
  author       = {Hongchuan Wei and
                  Wenjie Lu and
                  Pingping Zhu and
                  Guoquan Huang and
                  John J. Leonard and
                  Silvia Ferrari},
  title        = {Optimized visibility motion planning for target tracking and localization},
  booktitle    = {{IEEE/RSJ} Int. Conf. on Intel. Rob. \& Sys.},
  pages        = {76--82},
  year         = {2014},
}

@article{MCTH-sbmpamvut-05,
  author       = {Rafael Murrieta{-}Cid and
                  Benjam{\'{\i}}n Tovar and
                  Seth Hutchinson},
  title        = {A Sampling-Based Motion Planning Approach to Maintain Visibility of Unpredictable Targets},
  journal      = {Auton. Robots},
  volume       = {19},
  number       = {3},
  pages        = {285--300},
  year         = {2005},
}

@article{rl-mrtdtts-16,
  author       = {Cyril Robin and
                  Simon Lacroix},
  title        = {Multi-robot target detection and tracking: taxonomy and survey},
  journal      = {Auton. Rob.},
  volume       = {40},
  number       = {4},
  pages        = {729--760},
  year         = {2016},
}

@article{bmh-oplbnddvfvc-07,
  author       = {Sourabh Bhattacharya and
                  Rafael Murrieta{-}Cid and
                  Seth Hutchinson},
  title        = {Optimal Paths for Landmark-Based Navigation by Differential-Drive Vehicles With Field-of-View Constraints},
  journal      = {{IEEE} Trans. Rob.},
  volume       = {23},
  number       = {1},
  pages        = {47--59},
  year         = {2007},
}

@inproceedings{bh-enetppegvc-08,
  author       = {Sourabh Bhattacharya and
                  Seth Hutchinson},
  title        = {On the Existence of Nash Equilibrium for a Two Player Pursuit-Evasion Game with Visibility Constraints},
  booktitle    = {Algorithmic Foundation of Robotics},
  volume       = {57},
  pages        = {251--265},
  year         = {2008},
}

@article{ghnd-kmmtfvmc-11,
  author       = {Nicholas R. Gans and
                  Guoqiang Hu and
                  Kaushik Nagarajan and
                  Warren E. Dixon},
  title        = {Keeping Multiple Moving Targets in the Field of View of a Mobile Camera},
  journal      = {{IEEE} Trans. Robotics},
  volume       = {27},
  number       = {4},
  pages        = {822--828},
  year         = {2011},
}

@inproceedings{bah-mp3dttao-07,
  author       = {Tirthankar Bandyopadhyay and
                  Marcelo H. Ang and
                  David Hsu},
  title        = {Motion Planning for 3-D Target Tracking among Obstacles},
  booktitle    = {International Symposium Robotics Research},
  volume       = {66},
  pages        = {267--279},
  year         = {2007},
}

@book{w-vbopmp-17,
  author       = {Paul Keng{-}Chieh Wang},
  title        = {Visibility-based Optimal Path and Motion Planning},
  series       = {Studies in Computational Intelligence},
  volume       = {568},
  publisher    = {Springer},
  year         = {2015},
}

@article{kslo-prpp-96,
  author       = {Lydia E. Kavraki and
                  Petr Svestka and
                  Jean{-}Claude Latombe and
                  Mark H. Overmars},
  title        = {Probabilistic roadmaps for path planning in high-dimensional configuration
                  spaces},
  journal      = {{IEEE} Trans. Rob. Aut.},
  volume       = {12},
  number       = {4},
  pages        = {566--580},
  year         = {1996},
}

@TechReport{l-rrtnt-98,
  author        = {Steven M. Lavalle},
  title         = {Rapidly-Exploring Random Trees: A New Tool for Path
                  Planning},
  year          = {1998},
  institution   = {Iowa State University}
}

@article{ock-smpcr-24,
  author       = {Andreas Orthey and Constantinos Chamzas and Lydia
                  E. Kavraki},
  title        = {Sampling-Based Motion Planning: {A} Comparative
                  Review},
  journal      = {Annu. Rev. Control Robotics Auton. Syst.},
  volume       = {7},
  number       = {1},
  year         = {2024},
}

@inproceedings{lk-fmtvw-16,
  title     = {Follow Moving Things in a Virtual World},
  author    = {Lien, Jyh-Ming and Kim, Young J.},
  booktitle = {Korea Human-Computer Interaction Conf.},
  pages     = {69--73},
  year      = {2016}
}

@article{gchkskl-itamp-21,
  author       = {Caelan Reed Garrett and
                  Rohan Chitnis and
                  Rachel M. Holladay and
                  Beomjoon Kim and
                  Tom Silver and
                  Leslie Pack Kaelbling and
                  Tom{\'{a}}s Lozano{-}P{\'{e}}rez},
  title        = {Integrated Task and Motion Planning},
  journal      = {Annu. Rev. Control Rob. Aut. Sys.},
  volume       = {4},
  pages        = {265--293},
  year         = {2021},
}

@inproceedings{kl-rrtceasqpp-00,
  author       = {James J. Kuffner Jr. and
                  Steven M. LaValle},
  title        = {{RRT-Connect}: An Efficient Approach to Single-Query Path Planning},
  booktitle    = {{IEEE} Int. Conf. on Rob. \& Auto.},
  pages        = {995--1001},
  year         = {2000},
}

@inproceedings{smh-sbccrfo3de-05,
  author       = {Alejandro Sarmiento and
                  Rafael Murrieta{-}Cid and
                  Seth Hutchinson},
  title        = {A Sample-based Convex Cover for Rapidly Finding an Object in a 3-D Environment},
  booktitle    = {{IEEE} Int. Conf. on Rob. \& Auto.},
  pages        = {3486--3491},
  year         = {2005},
}

@article{sln-vbprmp-00,
  author       = {Thierry Sim{\'{e}}on and
                  Jean{-}Paul Laumond and
                  Carole Nissoux},
  title        = {Visibility-based probabilistic roadmaps for motion planning},
  journal      = {Adv. Robotics},
  volume       = {14},
  number       = {6},
  pages        = {477--493},
  year         = {2000},
}

@article{byam-lprrtdsrccf-22,
  author       = {Israel Becerra and
                  Heikel Yervilla{-}Herrera and
                  Emmanuel Antonio Cuevas and
                  Rafael Murrieta{-}Cid},
  title        = {On the Local Planners in the RRT* for Dynamical Systems and Their
                  Reusability for Compound Cost Functionals},
  journal      = {{IEEE} Trans. Robotics},
  volume       = {38},
  number       = {2},
  pages        = {887--905},
  year         = {2022},
}

@article{kslbh-pcrrtgkpfp-19,
  author       = {Michal Kleinbort and
                  Kiril Solovey and
                  Zakary Littlefield and
                  Kostas E. Bekris and
                  Dan Halperin},
  title        = {Probabilistic Completeness of {RRT} for Geometric and Kinodynamic
                  Planning With Forward Propagation},
  journal      = {{IEEE} Robotics Autom. Lett.},
  volume       = {4},
  number       = {2},
  pages        = {277--283},
  year         = {2019},
}


\end{document}